\pdfoutput=1

\documentclass[11pt]{article}

\usepackage[preprint]{acl}

\usepackage{times}
\usepackage{latexsym}

\usepackage[T1]{fontenc}

\usepackage[utf8]{inputenc}

\usepackage{microtype}

\usepackage{inconsolata}

\usepackage{graphicx}

\usepackage{float}
\usepackage{stfloats}
\usepackage{booktabs} 
\usepackage{tcolorbox}
\usepackage{listings}
\tcbuselibrary{listings,breakable}
\usepackage{amsmath,amssymb}

\usepackage{tikz}
\usetikzlibrary{positioning}

\title{DeMeVa at LeWiDi-2025: Modeling Perspectives with In-Context Learning and Label Distribution Learning}

\author{Daniil Ignatev, Nan Li, Hugh Mee Wong, Anh Dang, Shane Kaszefski Yaschuk \\
        Utrecht University, Utrecht, The Netherlands \\
        \{d.ignatev, n.li, h.m.wong, t.t.a.dang, s.p.kaszefskiyaschuk\}@uu.nl}

\DeclareMathOperator*{\argmax}{arg\,max}

\begin{document}
\maketitle

\begin{abstract}
This system paper presents the DeMeVa team's approaches to the third edition of the \emph{Learning with Disagreements} shared task \citep[LeWiDi 2025;][]{lewidi2025placeholder}.
We explore two directions: in-context learning (ICL) with large language models, where we compare example sampling strategies; and label distribution learning (LDL) methods with RoBERTa~\cite{liu2019-roberta}, where we evaluate several fine-tuning methods.
Our contributions are twofold: (1) we show that ICL can effectively predict annotator-specific annotations (\emph{perspectivist} annotations), and that aggregating these predictions into soft labels yields competitive performance; and (2) we argue that LDL methods are promising for soft label predictions and merit further exploration by the perspectivist community.

\end{abstract}

\section{Introduction}\label{sec:intro}
In natural language processing (NLP), annotations are often treated as a gold standard, implying a single, unambiguous truth.
However, for tasks that involve, among other things, cultural norms or subjectivity, human judgments can vary substantially, often reflecting diverse annotator backgrounds or personal perspectives \cite{plank-2022-problem,Cabitza_Campagner_Basile_2023}.
Customary approaches that aggregate these diverging annotations with techniques like majority voting disregard the 
potential validity of pluralistic interpretations, which may lead to the loss of valuable information about both the data instances and the people who annotated them.
The \emph{Learning with Disagreements} (LeWiDi) shared task shifts the focus to learning from unaggregated crowd labels, whether through learning from soft labels or through aligning models with specific annotators' viewpoints (i.e., \emph{perspectivist} training).

The DeMeVa team ranks 2nd overall on the leaderboard of the LeWiDi 3rd Edition shared task (LeWiDi 2025; \citealp{lewidi2025placeholder}). In this system paper, we describe the contributions of the DeMeVa team and discuss both our highest-scoring method and the other approaches that did not make it onto the leaderboard. We hope that our interpretation of these results will offer insights into learning with disagreement in NLP.

We obtained our score on the leaderboard by employing in-context learning (ICL) for perspectivist modeling.
ICL refers to the ability of pre-trained large language models (LLMs) to perform NLP tasks without task-specific training; in ICL, these models are instead conditioned on input-output examples (``demonstrations'') provided in the prompt \citep{brown2020-icl}.
Recent studies have demonstrated ICL's success on a wide range of tasks \cite[see e.g.][]{dong2024survey}. However, they have also shown that ICL is sensitive to the choice, order, and format of demonstrations. We explore how and to what extent ICL can be leveraged to steer LLMs toward the annotation patterns of individual annotators in natural language understanding.

In parallel with perspectivist ICL, our team also pursued alternative directions aimed at modeling label distributions. In this context, we drew on existing research from both NLP and other communities. Specifically, we refer to studies in \emph{label distribution learning} (LDL), a research vein that focuses on modeling probability distributions over full label spaces and which has its roots in the broader machine learning community.
We note that some of the insights from LDL have not yet fully found their way into NLP-specific research.
In our experiments, we build on such works by using two LDL-specific fine-tuning methods, neither of which has been widely applied in NLP: \emph{ordinal label distribution learning} \citep{wen2023-odl} and predicting population-level label distributions via clustering \citep{liu2019-population}.

The structure of this paper is as follows.
In Section~\ref{sec:data-and-evaluation}, we briefly reintroduce the datasets and sub-tasks of the LeWiDi shared task.
Next, we describe our ICL approaches in Section~\ref{sec:icl} and our LDL-related fine-tuning strategies in Section~\ref{sec:ft}.
Finally, we make our concluding remarks in Section~\ref{sec:discussion}.

\section{Datasets, tasks, and evaluation metrics}\label{sec:data-and-evaluation}
In this section, we discuss the datasets and evaluation metrics of the LeWiDi 2025 shared task.

\subsection{Datasets}\label{sec:datasets}

The LeWiDi 2025 shared task includes 4 datasets covering various aspects of natural language understanding (see Table~\ref{tab:datasets-overview} for an overview).

\begin{table*}[htb]
\centering
\resizebox{\textwidth}{!}{
\begin{tabular}{lcccc}
\toprule
\textbf{Dataset} & \textbf{Task} & \begin{tabular}[c]{@{}c@{}}\textbf{\#E} (train/dev/test)\end{tabular} & \textbf{\#Ann/E} & \textbf{\#Ann} \\
\midrule
CSC \citep{jang2024-csc} & Sarcasm detection & 5628/704/704 & 4+ & 840 \\
MP \citep{casola2024-mp} & Irony detection & 12017/3005/3756 & 5+ & 506 \\
Par (as yet unpublished) & Paraphrase detection & 400/50/50 & 4 & 4 \\
VariErrNLI \citep{weber2024-varierrnli} & NLI & 388/50/50 & 4 & 4 \\                
\bottomrule
\end{tabular}}
\caption{Overview of datasets used in LeWiDi 2025. \textbf{E} denotes entries, \textbf{Ann} denotes annotators.}
\label{tab:datasets-overview}
\end{table*}

\paragraph{CSC} The \emph{Conversational Sarcasm Corpus} \citep[CSC;][]{jang2024-csc} is a richly annotated sarcasm dataset containing 7,040 context-response pairs. For each of these pairs, the authors provided self-ratings on a 6-point Likert scale, and third-party annotators (360 in total, with 6 per author in Part 1 and 4 per response in Part 2) rated the level of sarcasm in the responses on the same scale.

\paragraph{MP} The \emph{MultiPICo Dataset} \citep[MP;][]{casola2024-mp} is a multilingual, socio-demographically grounded dataset of irony on social media, comprising 18,778 post-reply pairs from Reddit and Twitter across 9 languages and 25 linguistic varieties. Each received a mean of 5.02 binary irony labels from a pool of 506 crowd annotators balanced by gender and nationality.

\paragraph{Par} The \emph{Paraphrase Detection Dataset} (Par; as of yet unpublished) contains 500 sentence pairs from the \emph{Quora Question Pairs} dataset, each annotated by 4 expert annotators on a Likert scale ranging from -5 to +5 based on paraphrase quality.
Annotators were asked to provide short explanations justifying their scores as well.

\paragraph{VariErrNLI} The \emph{VariErrNLI} Dataset \citep{weber2024-varierrnli} is designed to disentangle genuine human label variation from annotation errors in natural language inference (NLI). It features a two-round annotation protocol applied to 500 multi-genre NLI (MNLI; \citealp{williams-etal-2018-broad, nie2020can}) items, resulting in 1,933 label-explanation pairs in the first round and 7,732 validity judgments in the second round. The dataset serves both as a benchmark for Automatic Error Detection methods and a resource to improve dataset trustworthiness. It also includes explanations for each annotation.

\subsection{Tasks and evaluation metrics}\label{sec:metrics}

LeWiDi 2025 introduces two tasks for the two established main approaches to unaggregated data: 1) Task A---\emph{soft label modeling}, where systems generate probability distributions over all classes for each item; and 2) Task B---\emph{perspectivist modeling}, where systems predict individual annotators' labels for specific items. At the same time, within each of these two tasks, the evaluation metrics vary depending on the structure of the concrete dataset they are paired with: e.g., Par and CSC, which both include Likert-scale values, require a different metric suite compared to datasets with unranked labels.

For Task A, the MP and VariErrNLI datasets make use of the Manhattan distance as the evaluation metric.
The Manhattan distance measures the sum of absolute differences between the predicted and the target distributions.
For VariErrNLI, this is extended to a \emph{Multi-label Average Manhattan Distance} (MAMD), averaging the Manhattan distances across multiple labels.
Performance on the Par and CSC datasets is assessed with the Wasserstein distance, which measures the minimum cost to transform one distribution into another.

Regarding Task B, MP and VariErrNLI are paired with the \emph{error rate} (ER) and \emph{multi-label error rate} (MER), respectively. ER measures the proportion of incorrectly matched values between predicted and target label vectors, while MER averages the error rates across multiple labels. For Par and CSC, the \emph{average normalized absolute distance} (ANAD) is used, which normalizes the average absolute difference between Likert scale values based on the range of the scale. In all cases, a lower score indicates better performance, with a score of 0 indicating a perfect match.

\section{In-context learning}\label{sec:icl}

Recent work has explored in-context learning for steering language models toward diverse human label distributions, primarily focusing on persona-based prediction for tasks like toxicity and hate speech detection \cite{sorensen2025value,10.1145/3477495.3531873,ramos-etal-2024-transparent}. In that vein, many studies focus solely on the effect of steering models with persona descriptions \cite{hu-collier-2024-quantifying,kambhatla2025beyond,sun-etal-2025-sociodemographic}; in the meantime, prompts that also incorporate annotations have been shown to elicit better predictions \cite{meister2025benchmarking}.
While these inquiries are mostly based on more widely used datasets, LeWiDi 2025 presents new challenges on tasks that have received limited attention so far in the domain of perspectivist NLP such as paraphrase evaluation and sarcasm detection.

We explore different ICL strategies on these novel datasets to advance perspective-aware modeling, leveraging state-of-the-art generative models: OpenAI's GPT-4o \cite{Achiam2023GPT4TR}, Claude Haiku 3.5 \cite{claude3modelcard}, and Llama 3.1 70B-Instruct \cite{grattafiori2024llama}. However, we do not explore persona-based steering as the LeWiDi 2025 datasets contain relatively few sociodemographic variables, making sociodemographic prompting infeasible.

\subsection{System pipeline}\label{sec:icl-pipeline}

\begin{figure}[htb]
  \centering
  \includegraphics[width=\linewidth]{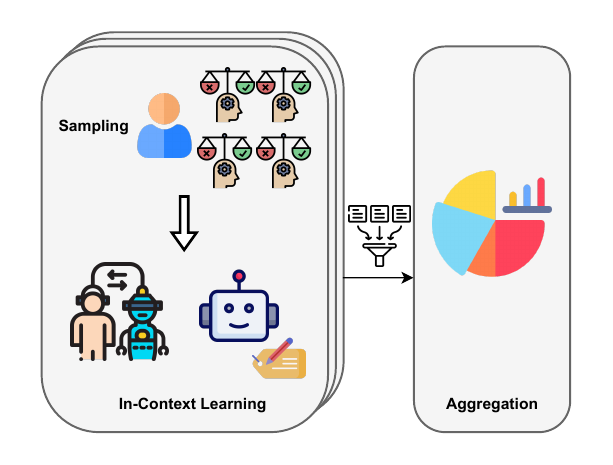}
  \caption{Our two-step pipeline to solve both tasks, based on ICL. In the first step (Task B), we sample examples from an annotator's past annotations and prompt the LLM to model annotator-specific behavior and predict labels for test inputs. In the second step, we aggregate these predictions into soft labels (Task A).}
  \label{fig:icl-pipeline}
\end{figure}

To accomplish both tasks of LeWiDi 2025, we propose a two-step pipeline (Figure~\ref{fig:icl-pipeline}).
First, we use ICL to prompt LLMs to predict individual annotators' labels based on their previous responses (Task B).
We then use these predictions to calculate the final soft label (Task A).

\begin{figure}[h!tb]
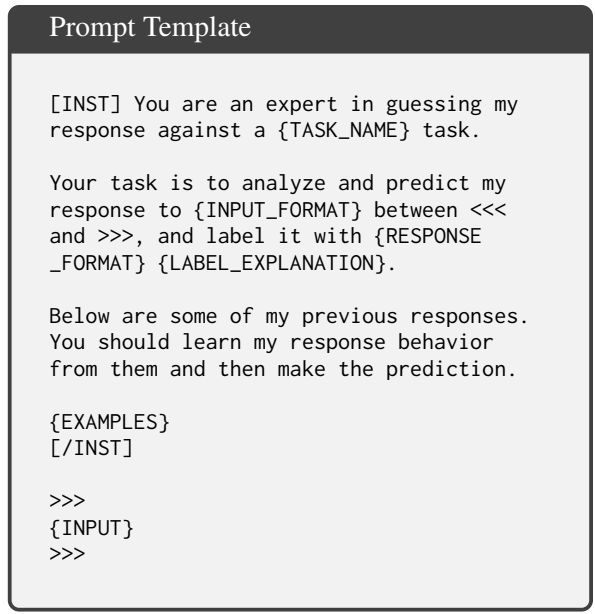

\centering
\begin{tcolorbox}[title=Prompt Template]
\begin{lstlisting}[basicstyle=\ttfamily\small\raggedright, breaklines=true, showstringspaces=false, columns=fullflexible, breakautoindent=false, breakindent=0pt]
[INST] You are an expert in guessing my response against a {TASK_NAME} task.

Your task is to analyze and predict my response to {INPUT_FORMAT} between <<< and >>>, and label it with {RESPONSE
_FORMAT} {LABEL_EXPLANATION}.

Below are some of my previous responses. You should learn my response behavior from them and then make the prediction.

{EXAMPLES}
[/INST]

>>>
{INPUT}
>>>
\end{lstlisting}
\end{tcolorbox}

\caption{Our ICL prompt template. The template supports varied tasks and input formats without sacrificing clarity.}
\label{fig:prompt-template}
\end{figure}

The two key components of ICL are \emph{demonstration selection} and \emph{prompt engineering}.
Our main focus is on finding the most appropriate example sampling method (demonstration selection).
As for the prompt engineering component, we use a simple template adapted from \citet{dutta2025annotator} that is applicable to all datasets in the shared task (see Figure~\ref{fig:prompt-template} for the prompt template and Appendix \ref{sec:prompt-example} for a filled example for the CSC dataset). The template is designed to be flexible enough to accommodate different tasks and input formats while also being straightforward enough for the LLM to leverage. For every experiment, we set the temperature to \texttt{0.0} to enforce greedy decoding and yield the most probable sequence with minimal randomness.

\subsection{Example selection strategies}

ICL is sensitive to how demonstrations are sampled and supplied to the model. We therefore compare two strategies for example selection: \emph{similarity-based} and \emph{stratified label-based} sampling. Additionally, we examine whether explanations available in the Par and VariErrNLI datasets can improve model personalization when included in prompts. This test builds on the work started by \citet{ye2022unreliability} and \citet{jiang-etal-2023-ecologically}, who stress the ambiguous role of explanations in NLI labeling.

The standard approach of retrieving semantically similar examples faces challenges with respect to perspectivist learning. BERT-based cosine similarity primarily ensures lexical and topical proximity \cite{kaster2021global}, but perspectivist tasks may require more nuanced selection. First, as \citet{10.1162/tacl_a_00523} show, annotators in NLU tasks rely on specific linguistic heuristics rather than topical similarity; hence, similarity with respect to these heuristics would offer a better selection criterion. Second, annotator-specific subsets can be arbitrarily small, which means they may lack enough similar examples for meaningful retrieval. Our two sampling strategies are as follows.

\paragraph{Similarity-based sampling}
For a test input $q$ (the current query) and annotator $a$, let $\mathcal{D}_a$ denote the set of training examples annotated by $a$. Let $\mathbf{h}(x)\in\mathbb{R}^d$ be the sentence embedding of $x$ produced by Sentence-Transformers \cite{reimers-2020-multilingual-sentence-bert}. We measure the relevance using the cosine similarity:
$\operatorname{s}(q,x) = \cos\big(\mathbf{h}(q),\,\mathbf{h}(x)\big)$.
We select $k$ demonstrations starting with $S=\varnothing$ and at each step, we add to this set the element
\[
  x^* \,=\, \argmax_{x\in \mathcal{D}_a\setminus S}\; \lambda\,\operatorname{s}(q,x)\; -\; (1{-}\lambda)\, \max_{x'\in S} \operatorname{s}(x,x')\textup{.}
\]

\noindent and update $S\leftarrow S\cup\{x^*\}$ until $|S|=k$. We set $\lambda=0.7$ to reduce redundancy among selected shots via the Maximal Marginal Relevance (MMR) method.

\paragraph{Stratified label-based sampling}
For each annotator $a$, let $\mathcal{D}_a$ denote the training set, $\mathcal{Y}_a$ the full set of their annotations, and $y_a(x) \in \mathcal{Y}_a$ the label assigned by this annotator for data sample $x$.
We first drop labels that occur less than two times to ensure stratification. Let $L=\max\{|\mathcal{Y}_a|,\,k\}$. If $|\mathcal{D}_a|\le L$ or only one label remains, we sample up to $k$ examples uniformly from $\mathcal{D}_a$.
Otherwise, we construct a stratified subsample $S'\subset\mathcal{D}_a$ that approximately preserves the empirical label proportions over $y_a(x)$. We do this using scikit-learn~\cite{scikit-learn}, and we then draw $k$ examples uniformly without replacement from $S'$.

We hypothesize that label-based sampling yields more representative examples by exposing models to diverse annotation patterns, which can be particularly effective for nuanced label scales (such as those found in CSC and Par) compared to binary tasks. This approach increases the likelihood that relevant linguistic heuristics appear in demonstrations, helping models learn annotator-specific decision patterns. We set the number of demonstrations to $k=10$.

\subsection{Model performance}

\begin{table*}[t]
\resizebox{\textwidth}{!}{
\begin{tabular}{l|cccc|cccc}
\toprule
 & \multicolumn{4}{c}{Task A} & \multicolumn{4}{c}{Task B} \\
 & CSC & MP & Par & VariErrNLI & CSC & MP & Par & VariErrNLI \\
\midrule
baseline\_random & 1.549 & 0.689 & 3.35 & 1.0 & 0.355 & 0.5 & 0.38 & 0.5 \\
baseline\_most\_frequent & 1.169 & 0.518 & 3.23 & 0.59 & 0.238 & 0.316 & 0.36 & 0.34 \\
\midrule
GPT-4o \textit{+sim} & 0.84 & 0.466 & 1.17 & 0.46 & 0.175 & 0.294 & 0.13 & 0.26 \\
GPT-4o \textit{+strat} & \textbf{0.792} & \textbf{0.469} & 1.25 & 0.44 & \textbf{0.172} & \textbf{0.3} & 0.14 & 0.25 \\
Haiku-3.5 \textit{+sim} & 1.005 & 0.657 & 1.58 & 0.43 & 0.205 & 0.375 & 0.15 & 0.26 \\
Haiku-3.5 \textit{+strat} & 0.95 & 0.684 & 1.47 & 0.42 & 0.201 & 0.392 & 0.16 & 0.27 \\
Llama-3.1-70B-Inst \textit{+sim} & 1.192 & 0.691 & 1.41 & 0.44 & 0.226 & 0.392 & 0.14 & 0.24 \\
Llama-3.1-70B-Inst \textit{+strat} & 1.157 & 0.706 & 1.38 & 0.36 & 0.227 & 0.399 & 0.15 & 0.22 \\
\midrule
\multicolumn{9}{l}{\textbf{\textit{+ Explanation}:}} \\
\midrule
GPT-4o \textit{+sim} & -- & --  & 1.17 & 0.43 & -- & -- & 0.12 & 0.24 \\
GPT-4o \textit{+strat} & -- & --  & \textbf{1.12} & \textbf{0.38} & -- & -- & \textbf{0.13} & \textbf{0.23} \\
Haiku-3.5 \textit{+sim} & -- & --  & 1.36 & 0.44 & -- & -- & 0.13 & 0.24 \\
Haiku-3.5 \textit{+strat} & -- & --  & 1.35 & 0.45 & -- & -- & 0.15 & 0.25 \\
Llama-3.1-70B-Inst \textit{+sim} & -- & --  & 1.35 & 0.46 & -- & -- & 0.14 & 0.25 \\
Llama-3.1-70B-Inst \textit{+strat} & -- & --  & 1.39 & 0.44 & -- & -- & 0.14 & 0.25 \\
\bottomrule
\end{tabular}
}
\caption{Results of ICL Strategies on LeWiDi 2025. \textit{+sim} denotes similarity-based example sampling, and \textit{+strat} denotes stratified label-based sampling. We additionally experiment with including annotator explanations in the Par and VariErrNLI datasets (\textit{+Explanation}). The results submitted to the leaderboard are shown in bold.}
\label{tab:results}
\end{table*}

We report the experiment results in Table~\ref{tab:results}.
While the performance differences between ICL approaches are relatively subtle, they mostly yield substantial improvements over the baseline methods across the datasets and tasks.\footnote{MP stands out as an exception to that. We explain the poor performance of Llama and Haiku on MP by the fact that they do not adequately support several of the languages present in MP.}

Similarity-based sampling performs best on MP, whereas label-based sampling tends to improve (lower) Task A distances on the other datasets without reducing the error rate.
For MP, both error rate and distance are lower when using similarity-based sampling. This is to be expected: with binary labels, stratified label-based sampling is practically equivalent to random sampling.
On the other three datasets, label-based sampling often results in improvements on Task A, while the error rate often changes insignificantly or even increases compared to stratified label-based sampling.
Our explanation for this is that the metrics for Task A show more sensitivity toward numeric values of predictions, and label-based sampling offers more control of said numeric values since the model limits its outputs to within the provided label range.
At the same time, since the error rate is not significantly influenced, our assumption that the sampled examples are more representative of this method does not appear to hold.

For Par and VariErrNLI, the results show that the inclusion of explanations further enhances performance; remarkably, this trend is more pronounced in Task A metrics compared to Task B metrics, as the error rate remains roughly in the same range for both settings. However, the calibration effect of label-based sampling is more notable (especially for GPT-4o), showing that it is amplified by reasoning examples. The fact that explanations improve performance in this regard may complement the results of \citet{ni2025reasoninghelplargelanguage}, who find that CoT-prompting helps steer RLHF models toward human perspectives. While explanations only contain one reasoning step, they still can be regarded as being analogous to more complex reasoning examples.

\subsection{Discussion}

To further validate the results of ICL, we examine its predictions on each development set in more detail. While the development sets of Par and VariErrNLI both comprise only 50 examples, those of MP and CSC consist of hundreds of items each, making it more challenging to inspect them thoroughly. We therefore sample a smaller subset of items from both datasets: for CSC, we randomly select 50 items, whereas for MP, we extract 50 random items for each included language (totaling 450 items). Although this strategy makes the analysis more feasible, it effectively prevents us from comparing per-annotator label distributions on CSC, as the down-sampled sets only include a few examples per worker. Nevertheless, we can still identify the strengths and weaknesses of our ICL methods based on specific data items.

\begin{figure}
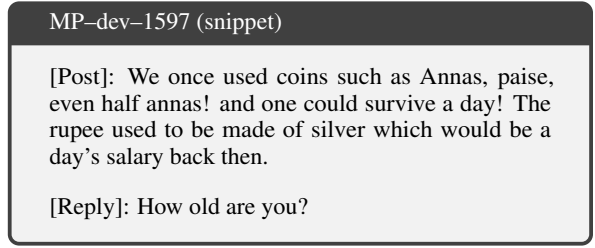

\centering
\begin{tcolorbox}[title=MP--dev--1597 (snippet), fonttitle=\small]
\small
\setlength{\parskip}{1em}   %

[Post]: We once used coins such as Annas, paise, even half annas! and one could survive a day! 
The rupee used to be made of silver which would be a day’s salary back then.

[Reply]: How old are you?
\end{tcolorbox}
\caption{A sample from the MP development set. The majority of the annotators marked this example as ironic.}\label{fig:mp_example}
\end{figure}

One notable tendency is that models often predict unanimous agreement on instances that appear straightforward on the surface, but which are actually annotated differently. We illustrate this with an example from \texttt{MP–dev–1597} in Figure~\ref{fig:mp_example}.
While the reply is directly licensed by the first utterance, that utterance does not give an immediate and obvious reason for an ironic reply. However, more than half of the annotators labeled this example as ironic. Similar cases can also be identified in the three other datasets, as annotators often demonstrate vastly different annotation behaviors. These examples possibly show that complete pluralistic alignment of language models may be impossible to fully achieve (at least in the linguistic domain), as the model adhering to common sense in all examples appears to be more important in this context when compared to adhering to the plurality of views.

At the same time, we note that the tested models are generally successful in mimicking specific annotators' labeling strategies. This is best illustrated in the VariErrNLI and Par datasets, since the annotators' motivations are directly available for analysis. For example, in the Par dataset, annotator \texttt{Ann3} uses label \texttt{0} considerably more often than their peers, consistently labeling most non-contradicting examples with $0$ rather than negative values even when they are non-relevant. This reasoning is also reflected in \texttt{Ann3}'s explanations. The models, particularly when combined with label-based sampling, tend to imitate this peculiarity while also never predicting $0$ for annotators \texttt{Ann1} and \texttt{Ann2} (who rarely use it). Likewise, the predictions also reflect less subtle differences, like the annotators' inclination toward positive or negative scale values (for Par) and entailment or neutral labels (for VariErrNLI). For example, \texttt{Ann3}'s preference for positive values is also discernible in the predicted labels. In this respect, it can be argued that ICL can be successful when used for perspectivist modeling of individual perspectives.

\section{Fine-tuning approaches}\label{sec:ft}
In this section, we discuss the various fine-tuning approaches we have explored for Task A.
While the overall performance of these approaches was ranked lower on the leaderboard than the in-context learning methods from Section~\ref{sec:icl}---in part because we merely tackled a subset of Task A---we believe that these fine-tuning methods are still a valuable contribution to understanding how we can learn from disagreements. All fine-tuning experiments were done using the base RoBERTa \citep{liu2019-roberta} model.

\subsection{Approach 1: Cumulative distances for Likert scales}\label{sec:ft-1}
In the machine learning and computer vision communities, \citet{geng2013-ldl} introduced \emph{label distribution learning} (LDL) as an alternative to \emph{single-label} and \emph{multi-label learning} (MLL).
While MLL allows data instances to be assigned to multiple classes, LDL aims to solve the ambiguity problem (i.e., instances potentially belonging to several classes) by predicting \emph{how much} each label describes an instance.
In other words, just like the soft evaluation approaches developed in perspectivist NLP communities, LDL predicts a probability distribution over the set of available labels.
\citet{wen2023-odl} remark that LDL algorithms generally fail to accurately predict distributions for tasks where the labels are inherently ordered, such as age estimation.
They propose the \emph{ordinal label distribution learning} (OLDL) paradigm and introduce evaluation metrics which take the ordinality of labels into account.

The Par and CSC datasets both contain annotations based on a Likert scale. These scales are ordered: higher ranks represent a higher degree of the measured concept.
In the first part of our fine-tuning efforts, we experiment with using two evaluation metrics proposed by \citet{wen2023-odl} as loss functions when fine-tuning RoBERTa: \emph{cumulative Jensen--Shannon divergence} and \emph{cumulative absolute distance}. During experimentation, we freeze all but the last six layers.

Let $\textup{CDF}_P$ and $\textup{CDF}_Q$ be the \textit{cumulative distribution functions}  of distributions $P$ and $Q$, respectively. We define the two loss functions as follows.

\paragraph{Cumulative Jensen--Shannon}
The \emph{cumulative Jensen--Shannon} (CJS) divergence between $P$ and $Q$ is defined as: 
\begin{equation}
    \textup{CJS}(P, Q) = \sum_{n=1}^C D_{js} (\textup{CDF}_P(n) || \textup{CDF}_Q(n))\textup{,}
\end{equation}
where $D_{js}(X || Y)$ denotes the Jensen--Shannon divergence between distributions $X$ and $Y$.

\paragraph{Cumulative Absolute Distance}
The \emph{cumulative absolute distance} (CAD) is defined as:
\begin{equation}\label{eq:cad}
    \textup{CAD}(P, Q) = \sum_{n=1}^C |\textup{CDF}_P (n) - \textup{CDF}_Q (n)|\textup{.}
\end{equation}

\noindent We make the following observation: the evaluation metric used for both Par and CSC in Task A is the Wasserstein distance (WSD).
Intuitively, the WSD reflects how much mass has to be moved, and how far, to transform one distribution into another.
In the discrete 1-dimensional scenario, as is the case for Likert labels, the Wasserstein distance reduces to:
\begin{equation}
    W_1(P, Q) = \sum_{n=1}^C |\textup{CDF}_P(n) - \textup{CDF}_Q(n)|\textup{,}
\end{equation}
which is the same as CAD (Equation~\ref{eq:cad}).
Indeed, \citet{wen2023-odl} proposed CAD as an adaptation of the Mallows distance, which is also known as the Wasserstein-2 distance.

\begin{table}[]
    \centering
    \begin{tabular}{l|cc}
    \toprule
        & CSC & Par \\ \midrule
         CJS & $0.831 \pm 0.01$ & $1.677 \pm 0.10$\\
         CJS+MAE & $0.813 \pm 0.00$ & $1.694 \pm 0.03$ \\
         CAD & $0.800 \pm 0.01$ & $1.590 \pm 0.12$\\
         CAD+MAE & $0.797 \pm 0.01$ & $1.558 \pm 0.10$ \\\bottomrule
    \end{tabular}
    \caption{Fine-tuning results (Wasserstein distance) for the CAD and CJS loss functions on the test sets. All results are averaged across three random seeds. Here, \texttt{CJS+MAE} is the average of predictions from CJS and MAE; \texttt{CAD+MAE} is defined in a similar fashion.}
    \label{tab:results_oldl}
\end{table}

\paragraph{Results}
We report our results in Table~\ref{tab:results_oldl}.
We concluded that straightforwardly using one of the given formulas as a loss function for fine-tuning RoBERTa would not be sufficient.
The reason for this is that although CAD is equal to the Wasserstein distance in 1D, minimizing CAD loss during training might not guarantee good generalization: the cumulative nature of CAD/W1 could allow small prediction errors to be diffused across subsequent labels.
As a result, we hypothesized that it might not strongly penalize localized prediction errors if the overall CDF stays close, potentially leading to blurry or smeared distributions.
For this reason, we also experimented with combining CJS/CAD with the mean absolute error (MAE), encouraging the mode of the predicted distribution to align better with the ``ground truth'' distribution while still respecting the ordinal structure of the data.
However, Table~\ref{tab:results_oldl} suggests that this does not make a difference.
For the CSC dataset in particular, we find that CAD/CAD+MAE can yield scores that are competitive with in-context learning (0.792 in Table~\ref{tab:results}).

\subsection{Approach 2: Population-level label distributions}

\citet{liu2019-population} introduce a strategy for learning label distributions designed to significantly reduce the total number of human labels required for each data item.
They suggest that even if humans can interpret a data item in many ways, their annotations tend to reduce these interpretations to a limited number of distinct ``ground truth'' label distributions.
Therefore, the annotations for any given item are seen as a sample drawn from one of these distinct underlying distributions.
They found that this technique works well for datasets with 5-10 annotations per data item. Given that the Par dataset only has four annotations per sentence pair, we used this approach on this dataset alone.
\citet{liu2019-population} also hypothesized that semantically similar items tend to have similar label distributions.
For this reason, they proposed to (1) cluster the data into semantically similar groups using unsupervised learning, (2) aggregate the annotations of the clusters to create a single soft label for each cluster, and (3) use supervised learning to learn to predict the unified label distributions.

When dealing with the Par dataset, we assume that some sentence pairs are inherently more difficult to annotate than others.
The annotations for these pairs may be more spread out and sparse as a result, while those for other samples may be more unified.
We adopt the clustering and two-stage training methodology proposed by \citet{liu2019-population}.
However, instead of using a single soft label distribution for all items in a cluster, we trained the classifiers on the original soft labels and then included clustering information to push the predicted soft labels to fall within a certain range.
 
\paragraph{Model Specification} For the clustering, we opted for $k$-means clustering with a maximum of 5 clusters. We clustered the sentence pairs into groups with similar soft label distributions and then used their cluster numbers to guide the training process.
We then fine-tuned RoBERTa to predict the soft labels.
To leverage the resulting clusters, we trained multitask classifiers with 2 prediction heads.

\paragraph{Soft Label Head} The soft label head is a simple feedforward layer outputting logits over 11 annotation scores from -5 to 5.
In this case, we used cross-entropy loss as the loss function. 

\paragraph{Cluster Classification Head}
To classify the clusters, we used a separate feedforward layer for predicting the logits for $n$ discrete cluster IDs. The head is trained to predict the corresponding cluster assignment of each example. For the loss function, we tried several options, namely KL divergence, Wasserstein distance, and all loss functions described in Section \ref{sec:ft-1}.

The overall training loss is the sum of the soft label loss and the weighted cluster classification loss:

\begin{equation}\label{eq:ltotal}
    L_{\textup{total}} = L_{\textup{soft}} + \alpha \cdot L_{\textup{cluster}}\textup{.}
\end{equation}

\noindent In this formula, $L_\textup{total}$ represents the total training loss, ${L_\textup{soft}}$ is the loss for soft label prediction, and
$L_\textup{cluster}$ is the loss for cluster prediction.
$\alpha$ is a tunable parameter that varies the overall influence of $L_\textup{cluster}$.

\paragraph{Results} Our best score with this approach is a Wasserstein distance of $1.66$ for the Par dataset. We achieved this by classifying the dataset into 3 clusters. While the performance is above the baseline by a notable margin, this method still underperformed compared to the other fine-tuning method described in Section \ref{sec:ft-1} and the in-context learning method described in Section \ref{sec:icl}.

\subsection{Discussion}
For the loss functions, it comes as no surprise that CAD yields better results than CJS on the test set, given that the evaluation metric is CAD/WSD.
Table~\ref{tab:results_oldl} suggests that a standard fine-tuning setup with this loss function might be enough to yield competitive scores on the CSC dataset.

Note that the Par dataset in particular had a relatively small number of annotators. Given that only four annotators were annotating on an 11-point Likert scale, sparse distributions are inevitable.
We find that our methods are not able to handle this sparsity well enough to yield scores comparable to those for CSC.
On a related note, we would like to make one additional observation:
when working with sparse annotations, it is highly important to consider how the models are evaluated.
When annotations are sparse, the ``ground truth'' distributions may only be a noisy, undersampled proxy of the underlying human opinion distribution.
As well, relying on raw empirical frequencies can exaggerate annotation noise, and evaluating against them with strict distance metrics such as the Wasserstein distance may unfairly penalize models that produce smoother (and arguably more plausible) distributions.
However, as it was not possible to apply smoothing to the unseen test set, we found that models optimized for smoother distributions will generally perform poorly according to the LeWiDi scoring mechanism.
All results reported in this section were obtained without additional smoothing.

As with many other domains, it appears that the NLP community can take inspiration from the computer vision and machine learning communities (and vice versa).
Indeed, the perspectivist approaches in NLP appear to have emerged independently from label distribution learning in CV/ML (and also with different objectives; note, for example, the fact that CAD and 1D WSD are the same), yet both grapple with similar challenges.
We argue that perspectivist NLP could benefit from the probabilistic and distributional modeling techniques developed in these other communities.

\section{Conclusion}\label{sec:discussion}
In this paper, we introduced the two main approaches taken by the DeMeVa team for the LeWiDi 2025 shared task.
Our comparison of ICL approaches on perspectivist modeling, while not yielding fully conclusive results, suggested that sampling examples based on labels can help generative models calibrate their predictions—especially for numeric outputs like Likert scale values.
Models calibrated in this way can trace and mimic annotators' behavior down to more specific, granular details.
However, their reliance on common sense (possibly induced by RLHF) may hinder their ability to recognize plurality when it is not overtly expressed.

The second contribution of this work is a call for the perspectivist NLP community to look outward.
In particular, we can learn from how machine learning communities have addressed uncertainty and label distribution learning.
While perspectivist NLP rightly centers the diversity of annotator perspectives, it can benefit from established techniques such as probabilistic modeling and smoothing methods that account for annotation noise and limited sample sizes.
We have merely scratched the surface here by borrowing simple loss functions and a clustering method from LDL, but we believe that engaging with other fields can be beneficial to the perspectivist community as a whole.

\section*{Ethical Considerations}
In this work, we make use of personalized annotations, which, \textit{inter alia}, include sociodemographic variables related to the annotators. However, anonymization by their respective original authors ensures that this data cannot be used in a manner that is harmful to individuals.

\section*{Acknowledgments}
We thank the anonymous reviewers for their helpful comments.
This work is funded by the Dutch Research Council (NWO) through the AiNed Fellowship Grant NGF.1607.22.002, \textit{Dealing with Meaning Variation in NLP}.

\bibliography{refs}

\newpage

\appendix
\section{Example of an ICL prompt}\label{sec:prompt-example}

\begin{figure}[H]
\centering
\begin{tcolorbox}[title=CSC--test--2143--\texttt{Ann743}, fonttitle=\small]
\begin{lstlisting}[basicstyle=\ttfamily\small, breaklines=true, showstringspaces=false, columns=fullflexible, breakautoindent=false, breakindent=0pt]
[INST] You are an expert in guessing my response against a sarcasm detection task.

Your task is to analyze and predict my response to a pair of context and response between <<< and >>>, and label it with an integer from 1 to 6 where 1 means not sarcastic at all and 6 means completely sarcastic.

Below are some of my previous responses. You should learn my response behavior from them and then make the prediction.

Example 0:
[Context]: Steve is a fan of Bulgarian folk music. Every week, he finds a different song and plays it on his phone and says, "I finally found one you'll like! This one is really good. Come on!"
[Response]: Bulgarian folk music is for old people Steve, didn't you say you wanted to be young and cool?
[Label]: 2

Example 1:
[Context]: You are watching TV with Steve. Whenever you set the volume to an odd number, Steve takes the remote control away from you and sets the volume to an even number.
[Response]: My mistake, I never useally do that.
[Label]: 2

...

Example 9:
[Context]: Steve and you are hanging out tonight. He shows up wearing a red tank top, green shorts, and yellow sneakers.
[Response]: Did you go to a yard sale or something?
[Label]: 5
[/INST]

>>>
[Context]: You walk into the room and Steve is wearing his shoes on his hands. When you see him, he says "look at me! I'm Mr. Shoehand!"
[Response]: Are you 5 or 50?
[Label]:
>>>
\end{lstlisting}
\end{tcolorbox}

\caption{ICL prompt for entry \texttt{CSC-test-2143} and Annotator \texttt{Ann743} (excerpt). The in-context examples are selected from \texttt{Ann743}'s annotations in the train set, following the stratified label-based sampling method.}
\label{fig:prompt-example-csc}
\end{figure}

\end{document}